\documentclass[conference]{IEEEtran}

\usepackage{amsfonts, gensymb, mathtools, microtype}

\usepackage[main=english, ngerman]{babel}
\usepackage[T1]{fontenc}
\usepackage[utf8]{inputenc}

\usepackage[hyphens]{url}
\usepackage[colorlinks, citecolor=LimeGreen]{hyperref}
\usepackage[dvipsnames, table]{xcolor}

\usepackage{tikz}
\usetikzlibrary{
    arrows,
    backgrounds,
    calc,
    decorations,
    calligraphy,
    fit,
    positioning,
    shapes,
    spy
}

\usepackage{pgfplots}
\pgfplotsset{compat=newest}
\usepgfplotslibrary{colorbrewer}

\usepackage[hang]{caption}

\usepackage{array, diagbox, makecell, multirow, subfig, xfrac}

\usepackage[export]{adjustbox}

\usepackage{graphicx, graphbox}
\usepackage[figuresright]{rotating}
\graphicspath{{figs/}}

\usepackage{listings}
\usepackage[defaultlines=3, all]{nowidow}

\newcolumntype{P}[1]{>{ \centering  \arraybackslash }p{#1}}
\newcolumntype{Q}[1]{>{ \raggedleft \arraybackslash }p{#1}}
\newcolumntype{M}[1]{>{ \centering  \arraybackslash }m{#1}}
\newcolumntype{N}[1]{>{ \raggedleft \arraybackslash }m{#1}}
\newcolumntype{B}[1]{>{ \centering  \arraybackslash }b{#1}}
\newcolumntype{C}[1]{>{ \raggedleft \arraybackslash }b{#1}}

\usepackage{ccicons}

\usepackage{fontawesome5}

\newcommand{\ORCID}[1]{\textsuperscript{\href{https://orcid.org/#1}{\textcolor[HTML]{A6CE39}{\faOrcid}}}}

\newcommand{\ORCIDSchlosser}{0000-0002-0682-4284} 
\newcommand{\ORCIDBeuth}{0000-0001-5482-9787}     
\newcommand{\ORCIDFriedrich}{0000-0001-6326-4749} 
\newcommand{\ORCIDKowerko}{0000-0002-4538-7814}   

\usepackage{fancyhdr, lastpage}
\fancypagestyle{plain}{\fancyfoot[C]{\thepage/\pageref{LastPage}}}
\begin{document}

\title{A Novel Visual Fault Detection and Classification System for Semiconductor Manufacturing Using Stacked Hybrid Convolutional Neural Networks}

\author{
    \IEEEauthorblockN{
        Tobias Schlosser\ORCID{\ORCIDSchlosser},
        Frederik Beuth\ORCID{\ORCIDBeuth},
        Michael Friedrich\ORCID{\ORCIDFriedrich}, and
        Danny Kowerko\ORCID{\ORCIDKowerko}
    }
    \IEEEauthorblockA{
        Junior Professorship of Media Computing, \\
        Chemnitz University of Technology, \\
        09107 Chemnitz, Germany, \\
        \texttt{\small \{firstname.lastname\}@cs.tu-chemnitz.de}
    }
}

\maketitle

\begin{abstract}
    Automated visual inspection in the semiconductor industry aims to detect and classify manufacturing defects utilizing modern image processing techniques. While an earliest possible detection of defect patterns allows quality control and automation of manufacturing chains, manufacturers benefit from an increased yield and reduced manufacturing costs. Since classical image processing systems are limited in their ability to detect novel defect patterns, and machine learning approaches often involve a tremendous amount of computational effort, this contribution introduces a novel deep neural network based hybrid approach. Unlike classical deep neural networks, a multi-stage system allows the detection and classification of the finest structures in pixel size within high-resolution imagery. Consisting of stacked hybrid convolutional neural networks (SH-CNN) and inspired by current approaches of visual attention, the realized system draws the focus over the level of detail from its structures to more task-relevant areas of interest. The results of our test environment show that the SH-CNN outperforms current approaches of learning-based automated visual inspection, whereas a distinction depending on the level of detail enables the elimination of defect patterns in earlier stages of the manufacturing process.
\end{abstract}


\section{Introduction and motivation}

Semiconductor manufacturing processes involve the development of models for capturing and monitoring manufacturing results. Hence, a manufacturing process incorporates a multitude of complex processing steps, ranging from the determination of the used material to the processing of the imaged circuits, whereas one of the steps is concerned with the cutting of the resulting components \cite{Huang2015}. The mechanical cutting and laser cutting processes used are characterized by a large number of parameters that influence the manufacturing process as a result of prevailing temperature, pressure, and voltage values. The nature of the resulting components ultimately provides information about the amount of flawless chips (yield) and occurring manufacturing defects. The aim is therefore the earliest possible detection of potential manufacturing defects, whereby manufacturers can benefit from an increase in yield and the reduction of manufacturing costs. Additionally, a manual inspection can mean a considerable and in particular exhausting time expenditure. Therefore, more and more visual inspection processes are automated.

The separation of silicon wafers into single components is called silicon wafer dicing, whereas the scribed regions of interest on the wafer surface are called dicing streets. As there exist various separation approaches, one of the more commonly deployed methods utilizes a dicing saw \cite{Hooper2015}. An alternative method is thermal laser separation, where a thermally induced mechanical force results in a cleave on the wafer surface \cite{Rahim2017}. In the case of thermally induced separation, the cleave is guided alongside the scribe. This process constitutes our quality criterion, as a cleave deviation can result in faulty chips and therefore a decrease in yield.

Our wafer data were obtained from real world dicing processes of different semiconductor wafers, as they were provided to us by a third party manufacturer. Initially, each wafer was mounted on a taped frame. After the cleaving process, the dicing tape was expanded to visualize the extremely thin cuts under a wide-field light microscope. The microscope scans each wafer line-wise, whereas the scanning stage allows, as in our application, imaging of up to 150~mm (6~inch) wide wafers with the option to stitch the resulting imagery.

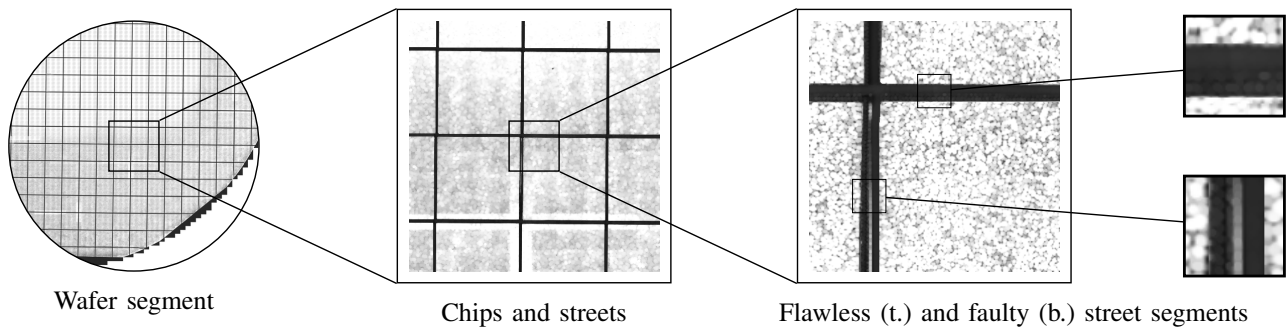
\begin{figure*}[tb]
    \centering

    \scalebox{1.33}{\begin{tikzpicture}[
     c/.style={draw, circle},
     r/.style={draw, rectangle},
     spy using outlines={rectangle, connect spies, magnification=3, size=1cm}]
        \node[c, minimum size=2.5cm, path picture={\node at (path picture bounding box.center)
         {\includegraphics[width=2.5cm]{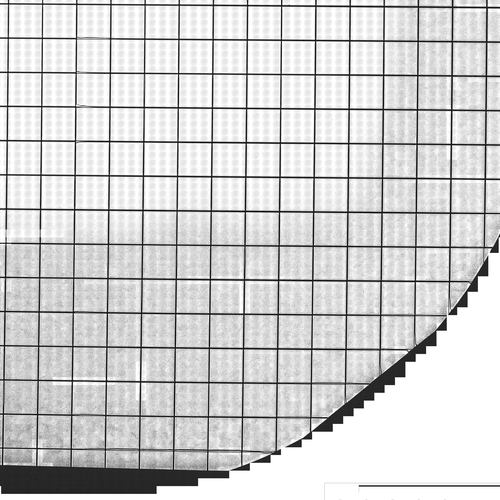}};}] (c) at (0, 0) {};

        \node[r, minimum size=0.5cm] (r1) at (c)    {};
        \node[r, minimum size=2.5cm] (r2) at (4, 0) {\includegraphics[width=2.5cm]{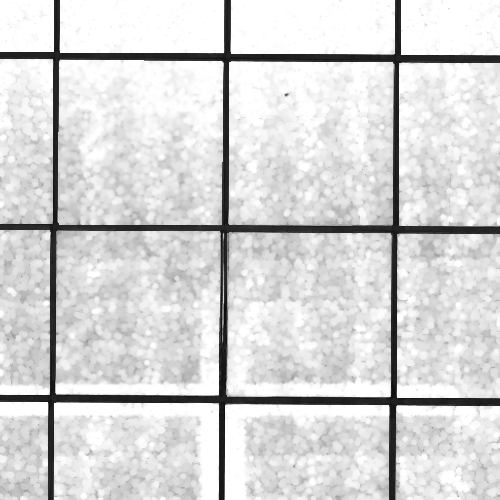}};
        \node[r, minimum size=0.5cm] (r3) at (r2)   {};
        \node[r, minimum size=2.5cm] (r4) at (8, 0) {\includegraphics[width=2.5cm]{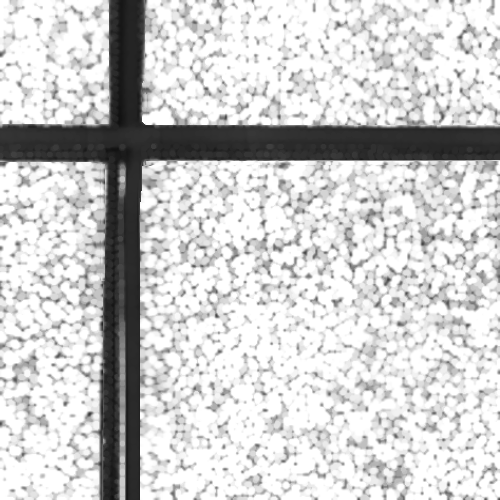}};

        \node[below=0.1cm of c,  scale=0.75] {Wafer segment};
        \node[below=0.1cm of r2, scale=0.75] {Chips and streets};
        \node[below=0.1cm of r4, scale=0.75] {\hspace{2cm} Flawless (t.) and faulty (b.) street segments};

        \draw (r1.north east) -- (r2.north west);
        \draw (r1.south east) -- (r2.south west);

        \draw (r3.north east) -- (r4.north west);
        \draw (r3.south east) -- (r4.south west);

        \coordinate (spypoint1)     at ( 8,     0.55);
        \coordinate (spypoint2)     at ( 7.35, -0.5);
        \coordinate (magnifyglass1) at (11,     0.8);
        \coordinate (magnifyglass2) at (11,    -0.8);

        \spy on (spypoint1) in node at (magnifyglass1);
        \spy on (spypoint2) in node at (magnifyglass2);
    \end{tikzpicture}}

    \caption{Wafer overview with chip and street segments.}
    \label{figure:wafer_overview}
\end{figure*}

Following, the detection and classification deals with manufacturing defects of wafer laser cuttings whose finest structures show often more complex defect patterns in pixel size and thus represent a particular challenge. Figure~\ref{figure:wafer_overview} shows this by means of a wafer segment (l.), subdivided into chip (m.) and street cuttings (r.), as they are produced through the cutting process alongside the scribes.

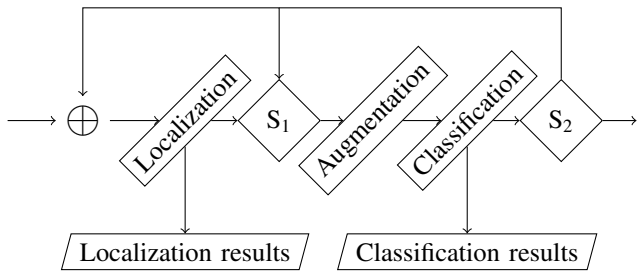
\begin{figure}[tb]
    \centering

    \begin{tikzpicture}[
     d/.style={draw, diamond},
     p/.style={draw, trapezium, trapezium left angle=75, trapezium right angle=105},
     r/.style={draw, rectangle, rotate=45}]
        \node[r] (s1) at (0,    0) {Localization};
        \node[d] (s2) at (1.25, 0) {S\textsubscript{1}};
        \node[r] (s3) at (2.5,  0) {Augmentation};
        \node[r] (s4) at (3.75, 0) {Classification};
        \node[d] (s5) at (5,    0) {S\textsubscript{2}};

        \node[   left  = 1cm   of s1.center] (s0)  {\LARGE $\oplus$};
        \node[p, below = 1.5cm of s1.center] (s1o) {Localization results};
        \node[p, below = 1.5cm of s4.center] (s4o) {Classification results};
        \coordinate[left  = 1cm   of s0.center] (s0i);
        \coordinate[above = 1.5cm of s2.center] (s2o);
        \coordinate[above = 1.5cm of s5.center] (s5o1);
        \coordinate[right = 1cm   of s5.center] (s5o2);

        \draw[->] (s1) -- (s2);
        \draw[->] (s2) -- (s3);
        \draw[->] (s3) -- (s4);
        \draw[->] (s4) -- (s5);

        \draw[->]  (s0i) -- (s0);
        \draw[->]  (s0)  -- (s1);
        \draw[->]  (s1)  -- (s1o);
        \draw[<->] (s2)  -- (s2o)  -| (s0);
        \draw[->]  (s4)  -- (s4o);
        \draw      (s5)  -- (s5o1) -- (s2o);
        \draw[->]  (s5)  -- (s5o2);
    \end{tikzpicture}

    \caption{Processing steps control flow graph of localization, augmentation, and classification with switches S\textsubscript{1} and S\textsubscript{2}.}
    \label{figure:processing_steps}
\end{figure}

\begin{figure}[tb]
    \centering

    \begin{tikzpicture}[r/.style={draw, rectangle, rotate=45}]
        \node[r]               (l1) at (0,   0) {Chip processing};
        \node[r]               (l2) at (1.5, 0) {Street processing};
        \node[r, align=center] (l3) at (3.5, 0) {Street-based \\ chip classification};

        \draw[->] (l1) -- (l2);
        \draw[->] (l2) -- (l3);
    \end{tikzpicture}

    \caption{Designed visual fault inspection system for chip and street localization, augmentation, and classification. Chip and street processing each undergo one iteration of the control flow graph shown in Fig.~\ref{figure:processing_steps}.}
    \label{figure:designed_system}
\end{figure}
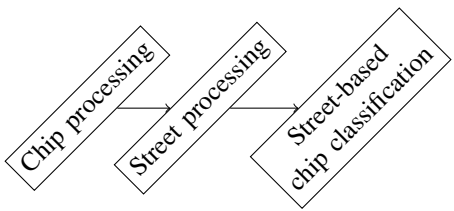

Classically, the field of automated visual inspection utilizes image processing approaches which are most commonly distinguished based on their functionality in projection-, filter-based, and hybrid approaches \cite{Huang2015}. Projection-based approaches often include principal component (PCA), linear discriminant (LDA), or independent component analysis (ICA), whereas filter-based approaches encompass spectral estimation and transformation-based approaches, including discrete cosine (DCT), Fourier (FT), and wavelet transforms. Furthermore, clustering approaches \cite{Huang2002} are often deployed as an additional classification step. However, since these approaches are also limited in their ability to detect novel defect patterns, they often result in a need for manual adaptation.

The first learning-based and hybrid approaches make use of multilayer perceptron (MLP) and support vector classifiers (SVC), as demonstrated by \textit{Chen and Liu} (2000) \cite{Chen2000}, \textit{Huang} (2007) \cite{Huang2007}, and \textit{Xie et al.} (2014) \cite{Xie2014}, including supervised and unsupervised approaches.

More recent research however demonstrates the benefits of deep neural networks. \textit{Lee et al.} (2017) \cite{Lee2017} deploy a one-dimensional time-expanded CNN that results in increased detection and classification rates. Even the latest research, following \textit{Nakazawa and Kulkarni} (2018) \cite{Nakazawa2018} and \textit{Cheon et al.} (2019) \cite{Cheon2019}, encompasses classical single CNN-based architectures. However, as these CNN-based approaches do not allow a distinction depending on the level of detail, an approach has to be found which enables the correction of defect patterns in the earlier stages of the manufacturing process.

In this contribution, we deal in contrast to that with the design of a multi-level hybrid system that combines the advantages of classical image processing approaches with artificial neural networks, thus allowing a more efficient detection of the finest structures in pixel size.

The core idea of our approach is inspired by findings from neuroscience, more namely the concept of visual attention. We observed that human workers pay in fact more attention to task-relevant regions of interest, thus focusing them. Such a process is in terms of human research or neuroscience known as visual attention. Visual attention has many findings, whereas a common one is the focusing of human processing on an aspect of a scene \cite{Hamker2005}. This can be like in our case a region, also known as spatial attention \cite{Carrasco2011}.

Several approaches exist to automatically identify these regions of interest and enhance their contribution for further analysis. We chose for applicability a classical computer vision pipeline. However, more complex and more neuroscience-related models are surely existing, e.g. saliency models \cite{Itti1998} or system-level attention models \cite{Hamker2005}.

\section{Fundamentals and implemented system}

Defects typically depend on the parameterization of the underlying cutting process or machine malfunctions, but also on human carelessness or exhaustion. Their nature and frequency of occurrence depend on the processing step in which they are produced. For the manufacturing process itself, a silicon wafer represents the base plate. The plate is then polished, covered with a layer of light-resistant material, and etched as a template of the subsequent circuit.

The existing error classes, such as small holes, broken out street corners, or misdirected laser cuts (see also Fig.~\ref{figure:wafer_overview}) indicate the complexity of inspecting variously good, erroneous, or conspicuous classes of patterns in a multitude of differently characterized circuits and structures. For this purpose, a distinction is made in the three classes of flawless, anomaly, and faulty patterns, each of which consist of a set of subclasses with corresponding frequency of occurrence and characteristics.

\subsection{Fundamentals}

In the following, we will introduce our stacked hybrid CNN-based approach, as it allows the localization of a region of interest (ROI). This is advantageous for the following CNN, enabling the localized ROI to be processed in a much higher resolution. The control flow graph in Fig.~\ref{figure:processing_steps} depicts the designed processing steps, divided into localization, augmentation, and classification.

The localization step realizes the focusing on a specific region. This step utilizes a set of classical image processing techniques to facilitate the applicability of the system, where a template of the integrated circuit layout has to be selected by the inspector to define the ROI. These regions are then localized depending on the defined template, cut, and output for following processing steps.

Afterwards, classical data augmentation is applied before the data is passed on to the CNN. For this purpose, depending on the level $l$ of the augmentation, an $l$-fold enhancement of the respective method takes place (notation $l\times, l \in \mathbb{N}_{\geq 0}$). This includes a rotation by $l \times \pm 2^{\circ}$, translation in $x$ and $y$ up to $l \times 5\%$, scaling up to $l \times \pm 2\%$, and reflection alongside the $x$ axis. The classification step is then realized via CNN, whereas the CNN architecture is chosen task-dependent on the specific application case.

\begin{table}[tb]
    \renewcommand{\arraystretch}{1.1}
    \centering

    \begin{tabular}{|M{1.25cm}|M{1.75cm}|M{2cm}|M{0.75cm}|M{0.75cm}|}
        \hline
        Unit & Layer & Type & Kernel size & Stride \\
        \hline
        \noalign{\vskip 2pt}

        \hline
        \multirow{3}{*}{$3 \times \text{conv}$} & conv1\_1 & convolution & $3 \times 3$ & $1$ \\
        \cline{2-5}
                             & conv1\_2 & convolution     & $3 \times 3$ & $1$ \\
        \cline{2-5}
                             & pool1    & max pooling     & $2 \times 2$ & $2$ \\
        \cline{2-5}
                             & dropout1 & dropout         & /            & / \\
        \hline
        \multirow{3}{*}{fc1} & dense1   & fully connected & /            & / \\
        \cline{2-5}
                             & dropout4 & dropout         & /            & / \\
        \cline{2-5}
                             & dense4   & fully connected & /            & / \\
        \hline
    \end{tabular}

    \caption{CNN layer configuration, consisting of three convolution blocks and one fully connected layer.}
    \label{table:model_configuration}
\end{table}

\subsection{Implementation of the system}

While the recognition of defect patterns in the finest structures in pixel size is dependent on the existing image resolution, a distinction is further made in the level of detail, whereby chips and streets are considered separately. Figure~\ref{figure:designed_system} represents the realized system and its processing steps.

\subsubsection{Chip processing}

If necessary, the wafer images are first separated into their chips. Afterwards, a localization of the separated chips takes place according to their position inside the wafer, i.e. the system separates the chips into inside and outside the wafer situated ones, including chips on the wafer border and beyond. In order to be able to counteract possible lower occurrences of individual error classes, we balanced all occurring classes. Afterwards, the chips are classified into inside and outside situated chips, as they are depending on the wafer border. This step is realized via convolutional neural network.

\subsubsection{Street processing}

The street regions are localized by employing a set of classical image processing techniques based on the template of the integrated circuit layout. At first, a histogram equalization is employed to enhance the recognition of edges. The enhanced image is then binary thresholded and an contour-based edge detection and border following is applied \cite{Suzuki1985}. To remove possible artifacts, an erosion serves as an additional processing step before the biggest contour is identified. On each side, the border center is determined and returned as the center of the street region.

At this point it may be noted, that the number of iterations, including chip and street processing, can be extended freely by providing additional templates.

The localized ROIs are directly used to detect and classify the respective error classes. Our realized CNN architecture configuration is inspired by \textit{Simonyan and Zisserman's} VGG network \cite{Simonyan2015} as shown in Table~\ref{table:model_configuration}.

\subsubsection{Street-based chip classification}

In the last step, the chip classification is returned based on the classified streets whose error classes are mapped according to their classified defect patterns, ranging from flawless to anomaly and faulty streets. Thus, for each chip, the street classification of its adjacent streets is adopted, while the degree of error is assigned as descending priority from faulty to flawless streets.

\begin{table}[tb]
    \renewcommand{\arraystretch}{1.1}
    \centering

    \begin{tabular}{|M{3cm}|M{2.5cm}|}
        \hline
        Test run & Mean accuracy \\
        \hline
        \noalign{\vskip 2pt}

        \hline
        RFC                    & $0.600 \pm 0.005$ \\
        \hline
        SVC with linear kernel & $0.677 \pm 0.001$ \\
        \hline
        SVC with RBF kernel    & $0.696 \pm 0.000$ \\
        \hline
        MLP                    & $0.681 \pm 0.020$ \\
        \hline
        \textbf{CNN}           & $\mathbf{0.757 \pm 0.032}$ \\
        \hline
        \noalign{\vskip 2pt}

        \hline
        \multicolumn{2}{|c|}{SH-CNN} \\
        \hline
        \noalign{\vskip 2pt}

        \hline
        $\mathbf{0\times}$ & $\mathbf{0.921 \pm 0.009}$ \\
        \hline
        $1\times$          & $0.896 \pm 0.013$ \\
        \hline
        $2\times$          & $0.909 \pm 0.011$ \\
        \hline
        $4\times$          & $0.880 \pm 0.022$ \\
        \hline
    \end{tabular}

    \caption{Street classification test results.}
    \label{table:test_results}
\end{table}

\section{Test results and evaluation}

In order to quantify the detection and classification capabilities of defect patterns of the realized system over chips and streets an evaluation of the individual iteration steps takes place. This includes the chip and street localization as well as the street augmentation (Fig.~\ref{figure:designed_system}), followed by the classification of the specific street and chip segments compared to current approaches of visual inspection.

For training itself, the machine learning frameworks Keras and TensorFlow were used, which also enable an accelerated processing by general-purpose graphics processing units. However, to allow a comparison with current learning-based approaches, the machine learning library scikit-learn was utilized.

The test run accuracies, as averaged over five runs for the selected approaches, random forest classifier (RFC), support vector classifier (SVC) with linear and RBF kernel, multilayer perceptron classifier (MLP), convolutional neural networks (CNN), as well as by means of the developed stacked hybrid CNN (SH-CNN), are shown in Table~\ref{table:test_results}. We conducted our test runs with a randomized data set split ratio of 50/25/25 for training, validation, and test set. Furthermore, the augmentation levels for deactivated ($0\times$) to high augmentation ($4\times$) are provided.

A comparison of all approaches reveals an evidently increased accuracy for the SH-CNN, whereby the augmentation level $0\times$ achieved the best street classification result. While the best accuracy of the previously tested approaches differ by about 17~\%, all SH-CNN test results show reduced differences. These can be explained in particular by the level of detail and the respective defect patterns, whereas the augmentation level itself enables the reduction of possible overfitting.

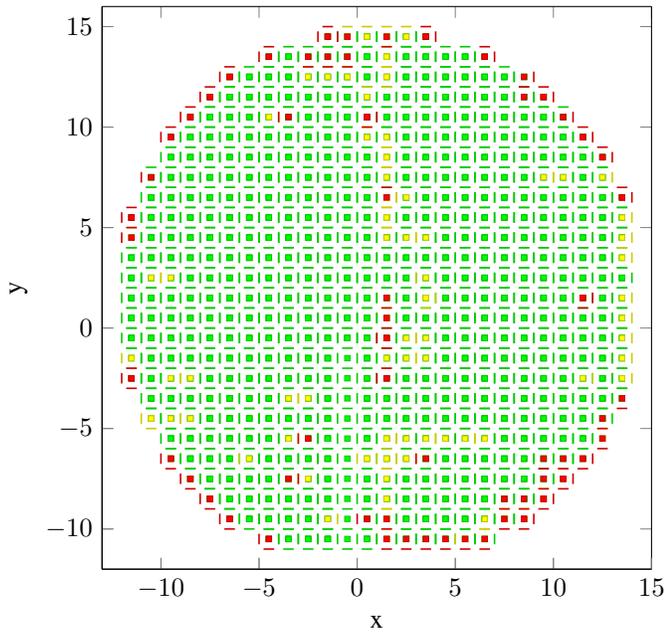
\begin{figure}[tb]
    \centering

    \begin{tikzpicture}
        \begin{axis}[
         colormap={summap}{color=(green); color=(yellow); color=(red)},
         mark size=1pt, point meta min=0, point meta max=2, view={0}{90},
         xmin=-13, xmax=15, ymin=-12, ymax=16,
         width=0.49\textwidth, height=0.5\textwidth, xlabel={x}, ylabel={y}]
            \pgfdeclareplotmark{hs}{\draw[semithick] (-2pt,  0pt) -- (2pt, 0pt);}
            \pgfdeclareplotmark{vs}{\draw[semithick] ( 0pt, -2pt) -- (0pt, 2pt);}

            \addplot3[scatter, x filter/.expression={x == round(x) ? nan : x}, mark=hs, only marks] table[col sep=comma] {data/streets.csv};
            \addplot3[scatter, y filter/.expression={y == round(y) ? nan : y}, mark=vs, only marks] table[col sep=comma] {data/streets.csv};

            \addplot3[scatter, mark=square*, only marks] table[col sep=comma] {data/chips.csv};
        \end{axis}
    \end{tikzpicture}

    \caption{Street and chip classification ground truth visualized for flawless (\textcolor{green}{\textbullet}), anomaly (\textcolor{yellow}{\textbullet}), and faulty (\textcolor{red}{\textbullet}) streets and chips.}
    \label{figure:wafer_visualization}
\end{figure}

The ground truth of the classified streets and resulting chip error classes is visualized in Fig.~\ref{figure:wafer_visualization}. This includes the street and chip classification test results for flawless (\textcolor{green}{\textbullet}), anomaly (\textcolor{yellow}{\textbullet}), and faulty (\textcolor{red}{\textbullet}) streets and chips according to the realized addressing scheme. The illustrated error classes can continue to differ in their respective defect pattern to be further assessed by an inspector.

\section{Summary and outlook}

The realized stacked hybrid convolutional neural network based (SH-CNN) system combines the advantages of classical image processing approaches with artificial neural networks and thus allows a more efficient recognition of the finest structures in pixel size. For this purpose, the in our application identified chip and street error classes constitute the base of the system, as they are provided to draw the focus over the level of detail from the wafer to its chips and streets. Our test results show that the realized system surpasses current approaches of learning-based automated visual inspection, whereby a distinction depending on the level of detail enables the detection and classification of defect patterns in the earlier stages of the manufacturing process.

Future projects can be built on top of this system and enhanced with, for example sensor-based analysis, such as audio or heat signatures. Furthermore, the system still has to be deployed under production test conditions, as the optic of the system, including camera and lightning settings, has to be investigated for further application-specific fine-tuning.

\section*{Acknowledgment}

This research is partially funded by the European Social Fund for Germany as well as the German Federal Ministry of Education and Research within the project group localizeIT.

\bibliographystyle{IEEEtran_Tobias}
\bibliography{library}

\end{document}